%% file: main.tex
\documentclass[a4paper,twoside]{article}

\usepackage[utf8]{inputenc}
\usepackage{lineno}
\usepackage{times}
\usepackage{graphicx}
\usepackage{booktabs}
\usepackage{float}
\usepackage{url}
\usepackage{color}
\usepackage{multirow}
\usepackage{mathrsfs}
\usepackage{dsfont}
\usepackage{epsfig}
\usepackage{subfigure}
\usepackage{calc}
\usepackage{amssymb}
\usepackage{amstext}
\usepackage{amsmath}
\usepackage{amsthm}
\usepackage{multicol}
\usepackage{pslatex}
\usepackage{apalike}
\usepackage{epstopdf}
\usepackage{array}
\usepackage{makecell}
\usepackage{SCITEPRESS} 

\begin{document}

\title{Real-Time Online Skeleton Extraction and Gesture Recognition on Pepper}

\author{\authorname{author 1, author 2}
}

\author{\authorname{Axel Lefrant\sup{1,2}, Jean-Marc Montanier\sup{1}}
\affiliation{\sup{1}Protolab, Softbank Robotics Europe, Paris, France}
\affiliation{\sup{2}INSA, University of Lyon, France}
\email{axel.lefrant@insa-lyon.fr, montanier.jeanmarc@softbankrobotics.com}
}


\abstract{\textit{We present a multi-stage pipeline for simple gesture recognition. The novelty of our approach is the association of different technologies, resulting in the first real-time system as of now to conjointly extract skeletons and recognise gesture on a Pepper robot. For this task, Pepper has been augmented with an embedded GPU for running deep CNNs and a fish-eye camera to capture whole scene interaction. We show in this article that real-case scenarios are challenging, and the state-of-the-art approaches hardly deal with unknown human gestures. We present here a way to handle such cases.
}}

\onecolumn \maketitle \normalsize \vfill
\input{sec_introduction.tex}
\input{sec_skeldetector.tex}
\input{sec_gesturereco.tex}
\input{sec_conclusion.tex}

\section*{ACKNOWLEDGEMENTS}

We express our most grateful acknowledgement to the members of the Protolab team of Softbank Robotics Europe (Maxime Busy, Maxime Caniot and Edouard Lagrue) without whom this work would not have been possible. We also thank Alexandre Mazel, the director of software innovation of Softbank Robotics Europe, for his technical knowledge and support on Pepper and the Jetson TX2.

\bibliographystyle{apalike}
{\small\bibliography{references-jm}}

\onecolumn{

\section*{\uppercase{Annexe 1}}

\begin{table}[H]
\centering
\begin{tabular}{c c c c c}
    & Vanilla & Vanilla neutral & Generate & Neutral \\
    \midrule
    \rotatebox{90}{Bowing} & 
    \thead{tpr: 0.65\\ uer: 0.15\\ usr: 0.08\\ fr: 0.05\\ dr: 0.06\\ tnr: 0.97\\ oer: 0.01\\ osr: 0.01\\ mr: 0.00\\ ir: 0.01} & 
    \thead{tpr: 0.70\\ uer: 0.12\\ usr: 0.07\\ fr: 0.05\\ dr: 0.05\\ tnr: 0.62\\ oer: 0.03\\ osr: 0.03\\ mr: 0.00\\ ir: 0.31} & 
    \thead{tpr: 0.52\\ uer: 0.20\\ usr: 0.09\\ fr: 0.04\\ dr: 0.14\\ tnr: 0.98\\ oer: 0.00\\ osr: 0.00\\ mr: 0.00\\ ir: 0.00} & 
    \thead{tpr: 0.61\\ uer: 0.17\\ usr: 0.10\\ fr: 0.51\\ dr: 0.65\\ tnr: 0.98\\ oer: 0.01\\ osr: 0.01\\ mr: 0.00\\ ir: 0.01}
     \\
    \rotatebox{90}{Clapping} & 
    \thead{tpr: 0.65\\ uer: 0.13\\ usr: 0.07\\ fr: 0.01\\ dr: 0.13\\ tnr: 0.97\\ oer: 0.01\\ osr: 0.01\\ mr: 0.00\\ ir: 0.02} & 
    \thead{tpr: 0.65\\ uer: 0.14\\ usr: 0.06\\ fr: 0.01\\ dr: 0.14\\ tnr: 0.97\\ oer: 0.00\\ osr: 0.01\\ mr: 0.00\\ ir: 0.02} & 
    \thead{tpr: 0.61\\ uer: 0.15\\ usr: 0.06\\ fr: 0.01\\ dr: 0.17\\ tnr: 0.98\\ oer: 0.01\\ osr: 0.00\\ mr: 0.00\\ ir: 0.00} & 
    \thead{tpr: 0.65\\ uer: 0.13\\ usr: 0.07\\ fr: 0.01\\ dr: 0.14\\ tnr: 0.98\\ oer: 0.01\\ osr: 0.00\\ mr: 0.00\\ ir: 0.01}
    \\
    {\rotatebox{90}{Drinking}} & 
    \thead{tpr: 0.73\\ uer: 0.11\\ usr: 0.08\\ fr: 0.03\\ dr: 0.05\\ tnr: 0.95\\ oer: 0.01\\ osr: 0.01\\ mr: 0.00\\ ir: 0.03} & 
    \thead{tpr: 0.74\\ uer: 0.10\\ usr: 0.09\\ fr: 0.02\\ dr: 0.06\\ tnr: 0.93\\ oer: 0.01\\ osr: 0.01\\ mr: 0.00\\ ir: 0.05} & 
    \thead{tpr: 0.64\\ uer: 0.14\\ usr: 0.09\\ fr: 0.03\\ dr: 0.09\\ tnr: 0.98\\ oer: 0.01\\ osr: 0.01\\ mr: 0.00\\ ir: 0.00} & 
    \thead{tpr: 0.72\\ uer: 0.11\\ usr: 0.09\\ fr: 0.02\\ dr: 0.06\\ tnr: 0.96\\ oer: 0.01\\ osr: 0.01\\ mr: 0.00\\ ir: 0.02}
    \\
    {\rotatebox{90}{Jumping}} & 
    \thead{tpr: 0.74\\ uer: 0.08\\ usr: 0.10\\ fr: 0.04\\ dr: 0.03\\ tnr: 0.97\\ oer: 0.01\\ osr: 0.01\\ mr: 0.00\\ ir: 0.02} & 
    \thead{tpr: 0.73\\ uer: 0.09\\ usr: 0.11\\ fr: 0.02\\ dr: 0.04\\ tnr: 0.97\\ oer: 0.01\\ osr: 0.01\\ mr: 0.00\\ ir: 0.02} & 
    \thead{tpr: 0.69\\ uer: 0.10\\ usr: 0.11\\ fr: 0.04\\ dr: 0.05\\ tnr: 0.98\\ oer: 0.01\\ osr: 0.00\\ mr: 0.00\\ ir: 0.01} & 
    \thead{tpr: 0.73\\ uer: 0.08\\ usr: 0.11\\ fr: 0.03\\ dr: 0.04\\ tnr: 0.97\\ oer: 0.01\\ osr: 0.01\\ mr: 0.00\\ ir: 0.01}
    \\
    {\rotatebox{90}{Waving}} & 
    \thead{tpr: 0.72\\ uer: 0.10\\ usr: 0.08\\ fr: 0.01\\ dr: 0.10\\ tnr: 0.98\\ oer: 0.01\\ osr: 0.00\\ mr: 0.00\\ ir: 0.01} & 
    \thead{tpr: 0.72\\ uer: 0.10\\ usr: 0.08\\ fr: 0.01\\ dr: 0.10\\ tnr: 0.98\\ oer: 0.00\\ osr: 0.00\\ mr: 0.00\\ ir: 0.01} & 
    \thead{tpr: 0.71\\ uer: 0.11\\ usr: 0.07\\ fr: 0.01\\ dr: 0.11\\ tnr: 0.99\\ oer: 0.01\\ osr: 0.00\\ mr: 0.00\\ ir: 0.00} & 
    \thead{tpr: 0.72\\ uer: 0.10\\ usr: 0.08\\ fr: 0.01\\ dr: 0.09\\ tnr: 0.99\\ oer: 0.01\\ osr: 0.00\\ mr: 0.00\\ ir: 0.01}
    \\
\end{tabular}
\caption{Action performance on all categories}
\end{table}
}

\end{document}

%% file: sec_introduction.tex
\section{\uppercase{introduction}}
\label{sec:introduction}

\begin{figure}
	\centering
	\includegraphics[scale=0.22]{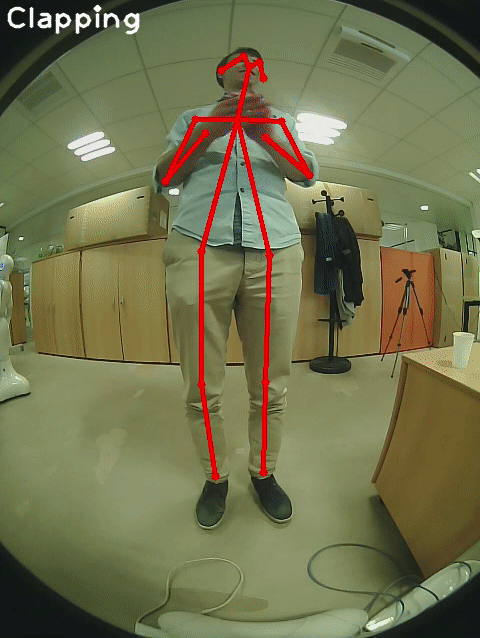}
	\caption{Example of the ``clapping" action.}
    \label{fig:waving}
\end{figure}

\noindent Pepper is an humanoid robot developed as a business welcoming and information gathering platform~\footnote{https://www.softbankrobotics.com/emea/en/robots/pepper}. To achieve his given tasks of helping people, Pepper has a basic face recognition system making it able to recognise emotions. However, communication with Pepper is currently limited to interacting with its on-board tablet. Desired use-case would be direct interaction with the robot. In this work, we enable Pepper to recognise simple gestures by analysing the skeleton of human extracted through the camera (Figure \ref{fig:waving}).

Deep learning techniques recently achieved outstanding result, especially in the computer vision field. Deep convolutional networks topped the major competitions like ImageNet \cite{imagenet} or COCO \cite{coco_dataset}. On top of that, deep networks improved robustness and accuracy in different sub-fields of computer vision (e.g. activity recognition, skeleton extraction) which were hardly feasible before. Hailed as an essential aspect of today's artificial intelligence, we investigate the potential and relevance of deep networks in the context of image recognition with mainstream robot like Pepper.

\bigskip
For the system to be intuitive and effective, it requires low-latency gesture detection and recognition. In the best case, the observational latency \cite{Ellis:2013:ETA:2440638.2440682}, that is the time needed to observe enough frames for an accurate decision, would be limited to the first frame of the gesture, which is unfeasible in practise. A reasonable performance latency would be a detection near the end of the gesture. To achieve this objective, we arbitrary decide on a system running at around $5$ frames per seconds, in order to sample enough frames for accurate detection at the end of the gesture while minimising the computational latency. The speed constraint guides us in the choice of a solution being fast and accurate.

\begin{figure}
	\centering
	\includegraphics[scale=0.2]{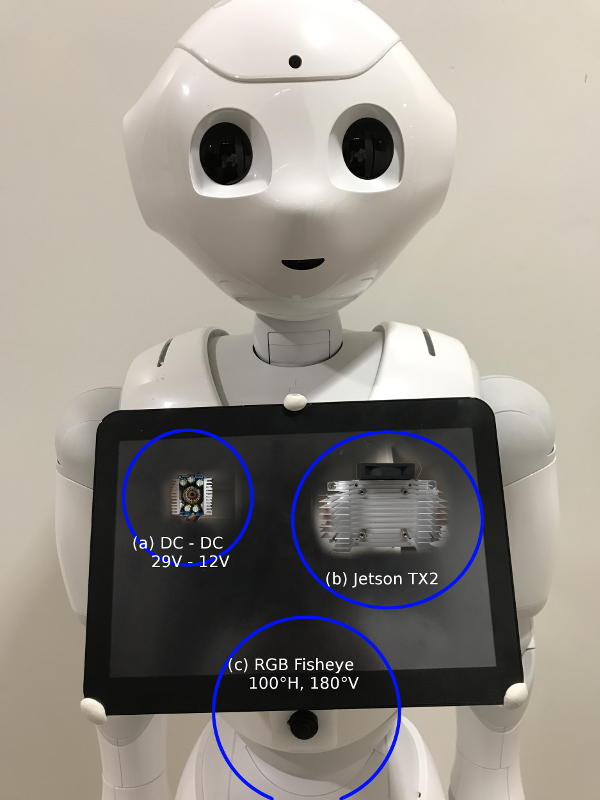}
	\caption{Augmented Pepper: (a) DC-DC converter to adapt Pepper's battery voltage to Jetson TX2 voltage; (b) Embedded Jetson TX2 board; (c) RGB Fish-eye rotated of $90^{\circ}$.}
    \label{fig:augmented_pepper}
\end{figure}

We aim to deploy the developed system on a Pepper robot. The current hardware of Pepper is insufficient for the task, which compel us to modify it (Figure \ref{fig:augmented_pepper}). Skeleton extractors based on convolutional networks consists of hundreds of convolutional layers, which cannot be run on the Pepper hardware. We opt for an embedded GPU of NVIDIA, the Jetson TX2, as this board is a good compromise between speed, memory and can be integrated inside a Pepper.

Also, both the central and mouth cameras have limited field of view, meaning that no adult usually fits entirely in the camera frame. Our solution is to include an additional fish-eye camera below the robot's Tablet, which allows both close and far interactions.


The developed system is composed of two hierarchical modules~\footnote{source code available at \url{removed for anonymisation}}, 1) a deep convolutional network extracting the skeletons of all humans body in a frame captured by a camera, and 2) a simultaneous detection of specific individuals and recognition of their gestures.

The originality of our approach is the association of different technologies, resulting in the first real-time system as of now to conjointly extract skeletons and recognise gesture on a Pepper robot. Moreover, we highlight a specific use-case due to the use of action recognition in the real-world: handling large portions of gestures and positions not known by the system. 

Section \ref{sec:skeldetector} discusses our work on the estimation of body-poses. Section \ref{sec:gesturereco} explains how the gesture are recognised. Section \ref{sec:neutral} focuses on handling the real-case scenario where unknown gestures are performed. Finally, Section \ref{sec:conclusion} concludes.

%% file: sec_skeldetector.tex
\section{\uppercase{skeleton detection}}
\label{sec:skeldetector}

\begin{table*}[t]
\centering
\caption{Evaluation of ArtTrack, Openpose and Mask R-CNN for body-pose estimation on COCO validation set.}
\resizebox{\textwidth}{!}{
\begin{tabular}{|c|l|l|l|l|l|l|}
\hline
 & \multicolumn{1}{c|}{\begin{tabular}[c]{@{}c@{}}Announced AP\\ (on test set)\end{tabular}} & \multicolumn{1}{c|}{\begin{tabular}[c]{@{}c@{}}Measured AP \\ (on val set)\end{tabular}} & \multicolumn{1}{c|}{\begin{tabular}[c]{@{}c@{}}Measured AR\\ (on val set)\end{tabular}} & \multicolumn{1}{c|}{\begin{tabular}[c]{@{}c@{}}Measured\\ time\end{tabular}} & \multicolumn{1}{c|}{\begin{tabular}[c]{@{}c@{}}CPU\\ Usage\end{tabular}} & \multicolumn{1}{c|}{\begin{tabular}[c]{@{}c@{}}GPU\\ Usage\end{tabular}} \\ \hline
ArtTrack \cite{arttrack} & / & 7.9 \% & 33.6 \% & \begin{tabular}[c]{@{}l@{}}637.78s \\ 128ms/frame (7.8 fps)\end{tabular} & \begin{tabular}[c]{@{}l@{}}25\% use\\ 2.0 GB memory\end{tabular} & \begin{tabular}[c]{@{}l@{}}30-80\% use (fluctuating)\\ 7.9 GB memory\end{tabular} \\ \hline
Openpose \cite{openpose} & $\sim$57 \% & 44.6 \% & 53.9 \% & \begin{tabular}[c]{@{}l@{}}611.8s\\ 122.4ms/frame (8.2 fps)\end{tabular} & \begin{tabular}[c]{@{}l@{}}50\% use\\ 1.2 GB memory\end{tabular} & \begin{tabular}[c]{@{}l@{}}90-95\% (stable)\\ 2.3 GB memory\end{tabular} \\ \hline
Mask R-CNN \cite{mask_rcnn} & 62.7 \% & 65.4 \% & 71.7 \% & \begin{tabular}[c]{@{}l@{}}1237.7s\\ 247.5 ms/frame (4.0 fps)\end{tabular} & \begin{tabular}[c]{@{}l@{}}33\% use\\ 1.6 GB memory\end{tabular} & \begin{tabular}[c]{@{}l@{}}40-80\% use (fluctuating)\\ 3.6 GB memory\end{tabular} \\ \hline
\end{tabular}}
\label{tab:baseline}
\end{table*}

\subsection{Related Works}

\subsubsection{Body-pose Estimation}
Following the results of single body-pose estimator \cite{deeppose}, \cite{confidence_heatmap}, \cite{cpm}, \cite{stacked_hourglass}, body-pose estimation was extended to multiple human pose detection. There are two different approaches for solving the multiple human pose estimation problem. 

The \textit{top-down} approach first detects persons in the image, crops the image around all the persons detected and finally performs human pose estimation on each cropped image \cite{Papandreou}, \cite{mask_rcnn}.

Opposed to the \textit{top-down} approach, the \textit{bottom-up} approach detects all the body parts in the image (location and label) and associates these parts to their respective human instance \cite{deepercut}, \cite{arttrack}, \cite{openpose}. 

The task of gesture recognition requires both accurate joint position even in the case of a distorted image (fish-eye camera) and a fast estimation to reliably capture the gesture sequence. Also, we would like this step to be as generic as possible and without prior assumption about the environment (e.g. location of people, number of people). To fulfil these requirements we are using multi-person 2D body-pose estimator networks.

\subsection{Evaluation Method}

In this section, we will evaluate which among the major and publicly available networks (i.e. Openpose \cite{openpose}, Mask-RCNN \cite{mask_rcnn} and ArtTrack \cite{arttrack}) are best suited for our application on the Jetson TX2.

\subsubsection{Dataset}

\noindent\textbf{Common Objects in Context (COCO) \cite{coco_dataset} ---}
The COCO dataset was created for the COCO challenge, including human body-pose estimation. This dataset contains around $200.000$ labelled images containing $1.7$ million labelled keypoints from over $150,000$ people instances.

The COCO dataset is using the \textit{Average Precision} (AP) for evaluating skeleton proposals. Each skeleton proposal is categorised as either a \textit{True-Positive} (TP) (there was a skeleton) and \textit{False Positive} (FP) (there was no skeleton), used to compute the AP measure:

\begin{equation}
\text{AP} = \frac{\vert \text{TP} \vert}{\vert \text{TP} \vert + \vert \text{FP} \vert}
\end{equation}

COCO defines the \textit{Object Keypoint Similarity} (OKS) to measure whether the skeleton proposal corresponds to a groundtruth skeleton. The OKS measure is adjusted by a threshold value. The final AP is an average over AP computed for different OKS threshold, from $0.5$ to $0.95$ with a stepsize of $0.05$.

\subsubsection{Implementation Details}

The three networks, Openpose, ArtTrack and Mask-RCNN, were trained on the COCO dataset. However, the test set annotations are not public. It appears that the validation set was not used to train the three networks, so we can safely use this set to evaluate the networks ($5000$ images in the validation set) without having to retrain the whole networks.

\subsection{Results}

\subsubsection{Baseline}
The results (Table \ref{tab:baseline}) were obtained on a personal computer, equipped with a Intel Xeon CPU and a NVIDIA GTX 1070 GPU. The measured AP on ArtTrack is surprisingly low, although the evaluation method is the same as the two other ones. Also, no result were reported on COCO dataset on the original paper (only on MPII Human Pose dataset \cite{mpii}), so no comparison is available. The other surprising result is the GPU memory usage, $7.9$ GB, which seems too much compared to the network architecture and to the other two networks. The authors were contacted, but no solution was found to explain or resolve this surprising result.

Concerning Openpose, the measured AP is slightly below what is announced in the paper. We only evaluate the `vanilla' Openpose, whereas the announced AP was obtained with additional multi-scale testing, post-processing with CPM \cite{cpm}, and filtering with an object detector. It is worth noting that the GPU usage of Openpose is kept almost constantly at top intensity, meaning that the API efficiently parallelises the workload.

Mask R-CNN displays the best accuracy but runs approximately two times slower than the others due to the multi-pipeline of the Faster R-CNN architecture. Looking at the CPU and GPU usage, we can also deduce that the underlying code can be optimised to further take advantage of the GPU (i.e. parallel image scaling and non-maxima suppression).

This experiment enables us to validate our experimental procedure and disregard ArtTrack due to its poor performance.

\subsubsection{Jetson TX2}
As the final system runs on the Jetson TX2 board, performance and time must be evaluated on it for a relevant choice. Given an equivalent image resolution, Openpose is faster while being less accurate than Mask R-CNN. From the two baselines, we draw the conclusion that Openpose is more suited for our application as we target a minimal framerate of about $5$ fps. Qualitative evaluation reveals that the skeleton detection is accurate enough for front face and close interaction with the robot.

\begin{table}[!ht]
\centering
\caption{Evaluation of Openpose and Mask R-CNN on Jetson TX2.}
\resizebox{\columnwidth}{!}{
\begin{tabular}{|c|l|l|l|}
\hline
 & \multicolumn{1}{c|}{\begin{tabular}[c]{@{}c@{}}Measured AP \\ on val set\end{tabular}} & \multicolumn{1}{c|}{\begin{tabular}[c]{@{}c@{}}Measured AR \\ on val set\end{tabular}} & \multicolumn{1}{c|}{\begin{tabular}[c]{@{}c@{}}Measured\\ time\end{tabular}} \\ \hline
\begin{tabular}[c]{@{}c@{}}Openpose\\ 320x176\end{tabular} & 22.7 \% & 30.5 \% & \begin{tabular}[c]{@{}l@{}}240.0 ms/frame \\ (4.2 fps)\end{tabular} \\ \hline
\begin{tabular}[c]{@{}c@{}}Mask R-CNN \\ -1x176\end{tabular} & 29.6 \% & 36.2 \% & \begin{tabular}[c]{@{}l@{}}587.7 ms/frame\\ (1.7 fps)\end{tabular} \\ \hline
\end{tabular}
}
\label{tab:baseline_tx2}
\end{table}

\subsection{Openpose Optimisation}

\subsubsection{TensorRT}
We now aim to further improve Openpose speed, by using the NVIDIA TensorRT technology which takes the embedded board capabilities into account. TensorRT is a deep learning framework, substituting traditional frameworks for optimised inference on NVIDIA platform. In addition to optimise model layout, TensorRT supports quantisation, which is the process of encoding the weight of the convolutional filter to lower-precision, compressing the size of the model, and usually accelerating inference since more operations can be performed. The Jetson TX2 only supports quantising the float on 16 bits (fp16), instead of the default encoding on 32 bits (fp32).

\subsubsection{TensorRT porting results}
We now evaluate Openpose optimised with the TensorRT framework on two different image resolution, $320 \times 240$ and $160 \times 112$, as the image resolution is a trade-off parameter between accuracy and speed. The results are presented in Tables~\ref{tab:tensort_320_240} and~\ref{tab:tensort_160_112}. In all conditions, the measured AP and AR stay the same when using TensorRT with or without quantisation. On the $320 \times 240$ resolution (Table~\ref{tab:tensort_320_240}), Openpose TensorRT fp32 is twice as fast as `vanilla' Openpose, and Openpose TensorRT fp16 is thrice as fast `vanilla' Openpose. On the $160 \times 112$ (Table~\ref{tab:tensort_160_112}), Openpose TensorRT fp16 is `only' twice as fast as the `vanilla' Openpose for the same resolution. Also, the performance of the detector with this resolution is too low to always accurately detect skeletons for front view interaction, especially with unusual body-pose. 

The difference in the factor of speed-up between the two resolutions is due to the time taken for network inference in the $160 \times 112$ resolution becomes similar to the time to group the keypoints to form human instances. As TensorRT only improves the network inference part, the speed-up becomes less significant. 

The final resolution used in the application is $320 \times 240$ pixels with fp16 quantisation because it allows us to accurately detect skeletons during front-view interaction while respecting our speed target. An additional advantage is the lack of rescale since the camera natively supports this resolution.

\begin{table}[!ht]
\centering
\caption{Evaluation of Openpose TensorRT for $320 \times 240$ resolution.}
\resizebox{\columnwidth}{!}{
\begin{tabular}{|c|l|l|l|}
\hline
 & \multicolumn{1}{c|}{\begin{tabular}[c]{@{}c@{}}Measured AP\\ on val set\end{tabular}} & \multicolumn{1}{c|}{\begin{tabular}[c]{@{}c@{}}Measured AR\\ on val set\end{tabular}} & \multicolumn{1}{c|}{\begin{tabular}[c]{@{}c@{}}Measured\\ time\end{tabular}} \\ \hline
\begin{tabular}[c]{@{}c@{}}Openpose Caffe\\ 320x240\end{tabular} & 30.6 \% & 39.2 \% & \begin{tabular}[c]{@{}l@{}}737.3 ms/frame \\ 1.4 fps\end{tabular} \\ \hline
\begin{tabular}[c]{@{}c@{}}Openpose TensorRT fp32 \\ 320x240\end{tabular} & 30.6 \% & 39.2 \% & \begin{tabular}[c]{@{}l@{}}347.9 ms/frame \\ 2.9 fps\end{tabular} \\ \hline
\begin{tabular}[c]{@{}c@{}}Openpose TensorRT fp16 \\ 320x240\end{tabular} & 30.9 \% & 39.1 \% & \begin{tabular}[c]{@{}l@{}}189.3 ms/frame \\ 5.3 fps\end{tabular} \\ \hline
\end{tabular}
}
\label{tab:tensort_320_240}
\end{table}

\begin{table}[!ht]
\centering
\caption{Evaluation of Openpose TensorRT for $160 \times 112$ resolution.}
\resizebox{\columnwidth}{!}{
\begin{tabular}{|c|l|l|l|}
\hline
 & \multicolumn{1}{c|}{\begin{tabular}[c]{@{}c@{}}Measured AP\\ on val set\end{tabular}} & \multicolumn{1}{c|}{\begin{tabular}[c]{@{}c@{}}Measured AR\\ on val set\end{tabular}} & \multicolumn{1}{c|}{\begin{tabular}[c]{@{}c@{}}Measured\\ time\end{tabular}} \\ \hline
\begin{tabular}[c]{@{}c@{}}Openpose Caffe \\ 160x112\end{tabular} & 10.4 \% & 16 \% & \begin{tabular}[c]{@{}l@{}}167.1 ms/frame\\ 6.0 fps\end{tabular} \\ \hline
\begin{tabular}[c]{@{}c@{}}Openpose TensorRT fp32\\ 160x112\end{tabular} & 10.4 \% & 16 \% & \begin{tabular}[c]{@{}l@{}}116.9 ms/frame\\ 8.55 fps\end{tabular} \\ \hline
\begin{tabular}[c]{@{}c@{}}Openpose TensorRT fp16\\ 160x112\end{tabular} & 10.4 \% & 16 \% & \begin{tabular}[c]{@{}l@{}}72.4 ms/frame\\ 13.8 fps\end{tabular} \\ \hline
\end{tabular}
}
\label{tab:tensort_160_112}
\end{table}

\subsection{Conclusion}

Thanks to state-of-the-art review and evaluation, we were able to find a network, i.e. Openpose, satisfying our speed constraint while enabling accurate skeleton detection. Further speed-ups with TensorRT make Openpose ample for our given task.

%% file: sec_gesturereco.tex
\section{\uppercase{Gesture recognition}}
\label{sec:gesturereco}

\noindent From the skeletons estimated by Openpose, we investigate on how to efficiently detect and recognise gestures.

\subsection{Related Works}

\subsubsection{Deep learning based recognition}

Deep learning has been successfully applied for skeleton-based gesture recognition, to automatically learn human representations from raw skeletal data. Both skeleton-based RNN \cite{du_hierarchical}, \cite{DBLP:journals/corr/ShahroudyLNW16}, \cite{attention_lstm} and CNN \cite{PHAM2018}, \cite{graph_skeleton} have been researched for gesture recognition, while RNNs have also been applied for online action detection from streaming skeleton data \cite{li2016online}, \cite{7952447}. While being promising, these networks are currently computationally expensive, especially on our GPU already filled with skeleton extraction.

\subsubsection{Skeletal features}

Contrary to deep networks that can implicitly learn representations, machine learning methods \cite{gesture_review_1}, \cite{gesture_review_2} require hand-crafted skeletal features that globally model the spatial structure and temporal dynamics of human skeleton. Common spatial features include displacement-based representations \cite{displacement_feature}, joint orientations features \cite{orientation_feature_1}, \cite{orientation_feature_3}. Popular temporal features are the velocity and the acceleration of the joints and angles since they encode quite well motions. Fusing these different modalities allow space-time human representations.

The previous spatial and temporal features are considered low-level, since they are directly extracted from the skeleton data in euclidean space. Additional human representations can also be constructed from feature transition to other topological space, i.e. manifold representations \cite{lie_manifold}, \cite{grassmann_manifold}. Another mid-level features are the body-parts representations that partially take into account the physical structure of the human body \cite{Evangelidis-ECCVW-2014}, \cite{7899766}.

\subsubsection{Gesture classification}

One approach is to consider action recognition as a time series problem in which the characteristics of body postures and their dynamics over time are extracted to represent a human action. Hiden Markov Models (HMMs) were extensively used in the field of recognition, e.g., hand gesture recognition \cite{Yin2014RealtimeCG}, \cite{6566433}. The advantage of HMM is its ability to learn the temporal dependencies of the gesture from a limited training set. However, HMMs are inherently only able to model dynamic movement and ignore static poses.

The template-based methods treat gesture and action recognition as a database query problem which matches data with templates in the database \cite{Evangelidis-ECCVW-2014}, \cite{6376770}. The intuitive idea in the bag of features is to generate a code-book of features, and represent each video with an histogram. The visual codebook is generally built by clustering descriptors extracted from the objects of interest, the centers of the clusters are considered the visual words. Online detection and recognition has been achieved using the template-based method \cite{6376770}, \cite{DBLP:journals/corr/MeshryHT15}.

\subsection{Online Gesture Detection and Recognition}
The method selected for the gesture recognition is a template-based method. Specifically, we decided to implement the algorithm proposed by \cite{DBLP:journals/corr/MeshryHT15} because of its performances and applicability to real-time scenarios.

\subsubsection{Algorithm}
The algorithm is learning from sequence manually annotated and representing the demonstration of an individual performing a single action or gesture, from start to finish. The training set is composed of $S$ sequences, and each sequence is composed of $N$ frames. Given a sequence of $N$ set of body-joints representing the successive joints position when realising a specific action, meaningful features are extracted at each frame. The features are composed of the Moving Pose \cite{moving_pose} features and angle features \cite{shotton}. The final form of the descriptor, called gesturelets, is as follow:

\begin{equation}
g = [P(t), \alpha \frac{P(t)}{dt}, \beta \frac{P^2(t)}{dt^2}] \gamma [ \frac{\Theta(t)}{dt}]
\end{equation}

where $\alpha$, $\beta$ and $\gamma$ are hyper-parameters (detailed in section \ref{sec:hyperparameters}) weighting the contribution of each feature, and $P(t)$, $\frac{P(t)}{dt}$, $\frac{P^2(t)}{dt^2}$ and $\frac{\Theta(t)}{dt}$ respectively refer to relative joint position from hip center, joint speed, joint acceleration and joint angle speed. This descriptor is normalised to unit norm for scale invariance. 

We cluster the descriptors from all the sequence of the training set to form a codebook of $K$ centroid. Each action sequence is represented as an histogram, of size $K$, which counts how many times each cluster has been assigned to each gesturelets of the sequence. Soft binning can be applied to adjust the cluster importance depending on its distance from the gesturelet when performing k-nearest neighbours. 

From the $S$ histograms (one per sequence), a multi-class linear SVM classifier is trained in a \textit{one-vs-all} approach, although other linear classifier can be used. From this classifier, we only retain the weight vector $w$, which characterises the separating hyperplane, i.e. how much the clusters are important in regard to their related gesture.

Given a sequence of $N$ frames, we extract a descriptor for each frame. Then for each of these descriptors, we find its k-nearest neighbours with the clusters of the codebook. We compute an action score for each mapping by looking at the weight vector of the action. A detected action is triggered when the score of the maximum subarray, computed in linear time with Kadane’s algorithm \cite{kadane_algorithm}, exceeds a threshold. This threshold is determined by searching for the score threshold that minimises the binary classification error for its related action class.

The online procedure is identical, except that the sequence is coming frame by frame, so we continuously perform maximum subarray search until an action is detected.

\subsubsection{Experimentation}
A missing element of the original detector is a probability score measuring the relevance of the detection and confidence on the predicted action class. To compute a probability score, we store the cluster matching of each descriptor and when a gesture is triggered, we compute the histogram of the sequence from the start and end value deduced by the detector. Then this histogram is classified by the SVM, alongside a calibrated logistic regression giving a probability score. We call this implementation `\textit{vanilla}'.

A particularity of the method developed is to decide when a detection is triggered. By default, a detection is triggered only when an action threshold is overtaken. Another way is to also require that the following $s$ consecutive scores are negatives. The idea is to avoid short detection or multiple same detection for one gesture. This is done at the expense of the observational latency, since it should force the system to wait until the end of the gesture. However, depending on the dataset used for learning, this method may inhibit some gestures since the start and end of the gesture may be similar. To overcome this problem, the authors of~\cite{DBLP:journals/corr/MeshryHT15} proposed a mixing histogram scheme. We call this solution `\textit{generated}'.

\subsection{Dataset}

\noindent\textbf{NTU RGB-D \cite{DBLP:journals/corr/ShahroudyLNW16} ---}
This action recognition dataset consists of $56.880$ samples of $60$ different actions containing RGB videos, depth map sequences, 3D skeletal data, and infrared videos for each sample. In our case, only the 3D skeletal data is necessary for training the classifier. The dataset is captured by $3$ Microsoft Kinect v.2 cameras concurrently, capturing accurate 3D skeletal data of $25$ major body joints, at each frame and from different point of view (front view, +$45^\circ$, -$45^\circ$). \smallskip

\noindent\textbf{SBRE neutral poses and detection ---}
When interacting with a robot, the user often stays motionless in a pose that could be qualified as `\textit{neutral}'. This category of movement is particularly important to us, but not present in the original NTU RGB-D dataset. We have therefore generated a new dataset to fulfil our needs. It consists of samples of individuals standing clearly in front of the fish-eye camera equipped by Pepper. When capturing the set, Pepper is in its standing pose, and the subject is positioned at either close (0.5m), medium (1.5m) and far (3m) distance from the camera. The videos are recorded at 30fps with the resolution of $640 \times 480$ pixels.


Cleaning of the NTU RGB-D dataset has been done to remove some variability in the action demonstration, enabling the SVM classifier to better learn the action representations. The SBRE neutral dataset as well as the list of files sampled from the NTU RGB-D dataset can be freely downloaded~\footnote{\url{removed for anonymisation}}.

\subsection{Results}

\subsubsection{Hyper-parameters optimisation}
\label{sec:hyperparameters}

The detector is composed of six hyper-parameters, specifically three for features extraction ($\alpha, \beta, \gamma$), one controlling the size of the codebook (i.e. number of cluster), one for the number of cluster neighbour in the soft binning and one for how much we weight the positive scores against the negative scores when we learn the threshold. The search for good values of these hyper-parameters is done with cross-validation on the recognition task only. Our assumption is that the hyper-parameters found in these conditions are acceptable for the detection task on a fish-eye camera. 

The cross-validation result (Table \ref{tab:hyperparameters}) emphasises the angle features, while the joint acceleration seems to have less influence in the subsequent classification. 

\begin{table}[!ht]
\centering
\caption{Best hyper-parameters found with coarse-grain cross-validation.}
\resizebox{\columnwidth}{!}{
\begin{tabular}{|c|c|c|c|c|c|c|}
\hline
$\alpha$ & $\beta$ & $\gamma$ & \begin{tabular}[c]{@{}c@{}}codebook\\ size\end{tabular} & \begin{tabular}[c]{@{}c@{}}m\\ neighbours\end{tabular} & \begin{tabular}[c]{@{}c@{}}weight\\ factor\end{tabular} & \begin{tabular}[c]{@{}c@{}} recognition\\ score \end{tabular} \\ \hline
0.8 & 0.4 & 1 & 2000 & 2 & 3 & 0.888 \\ \hline
\end{tabular}
}
\label{tab:hyperparameters}
\end{table}

\subsubsection{Evaluation metric for gesture detection}

One challenge is to effectively measure the detection performance of our detector. We divide the NTU RGB-D dataset of pre-segmented clips into a training set and a test set, and we randomly concatenate the sequences of the test set to effectively make long sequences of multiple actions. While this trick allows us to evaluate the performance of the detector, the final score will not be totally representative of the performance of the detector in practical application (e.g. fish-eye distortions, noisy skeletons). 

Also, the metric and way to measure the detection are essentials. A simple solution is to rely on precision and recall measures and finally compute a F1 score. While this approach is often used in other works and make comparisons straightforward, the details needed for its computation is generally not provided (e.g. are the computation based on the frames or events) and it usually does not offer much insight on the quality and failure of the detector. 
We adopt the metrics designed specifically for activity recognition \cite{metrics_activity_reco}. This evaluation procedure introduces a scoring segments approach, making possible binary comparison between the groundtruth and prediction for each action. Besides the regular measures of true-positives, true-negatives, false-positives and false-negatives, additional metrics are designed to provide better information on the reason of failure (such as fragmentation, deletion or insertion). Finally, the evaluation's computation is fully detailed which allows easy reproduction.

\subsubsection{Original implementation evaluation}

We evaluate first the `\textit{vanilla}' and `\textit{generated}' approaches with a set of 5 actions to recognise, `bowing', `clapping', `drinking', `jumping', and `waving'. These particular actions were chosen because they use almost all limb, while being globally different. In total, after cleaning, the training set contains $1091$ sequences for a total of around $12000$ frames. 

For each condition, training has been done 100 times, and each time one evaluation has been performed. As proposed in~\cite{metrics_activity_reco}, for each action, all the measures can be gathered in two groups: one showing the results obtained in the positive cases (when the class to detect is present) and the one showing the negative cases (when the class to detect is not present). Each measure represents the average over all training performed. 

The Figure~\ref{fig:vanilla-generated-neutral} presents these data as pie charts for the `bowing' action. This action has been chosen because it is the hardest to classify in our set, and therefore the one that shows more of the classifiers' performances. The analysis of a `NULL' action are also presented: these corresponds to lack of output (i.e. no detection) when a gesture is presented to the classifier. All results for all actions are available in Appendix 1.

The first line presents the analysis of the classification of the `bowing' action with the `\textit{vanilla}' approach. The positive and negative rates have been computed by taking into account only the 5 actions the system has been trained on. In the second line we include `neutral' actions (from the SBRE Neutral dataset) to the test. One can see that with this additional action in the test set, the true negative detection rates ($tnr$) drops from 0.97\% to 0.63\%, meaning the `\textit{vanilla}' approach has difficulty handling real-case scenarios.

The third line presents the results obtained with the `\textit{generated}' approach when tested on the 5 actions used for training plus the neutral actions. It appears that the \textit{`generated'} approach brings back the rate of true negative detections ($tnr$) close to the one obtained with the \textit{`vanilla'} approach. However this solution comes at the cost of a large reduction of the true positive detection rate ($tpr$) from 0.653\% to 0.527\%. Moreover, we can see that with the \textit{`generated'} approach the $tnr$ is also strongly reduced when neutral poses are presented. The fact that the insertion rate ($ir$) has increased indicates that the detector fired-off less often.

\begin{table*}[t]
\centering
\caption{Bowing performance on all categories}
\resizebox{\textwidth}{!}{
\begin{tabular}{c c c c}
    & Bowing Negative & Bowing Positive & NULL Negative \\
    \midrule
    \raisebox{2\normalbaselineskip}[0pt][0pt]{\rotatebox{90}{COCO dataset}} &
    \includegraphics[width=0.3333\textwidth]{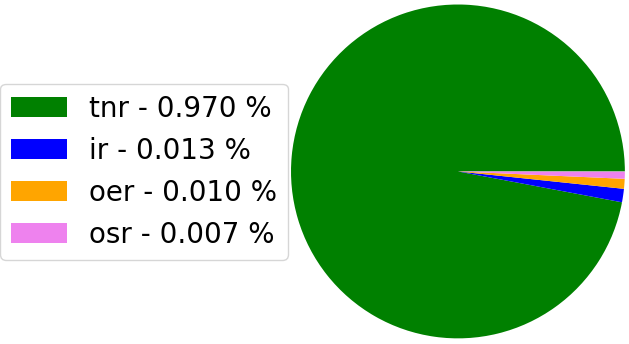} &
    \includegraphics[width=0.3333\textwidth]{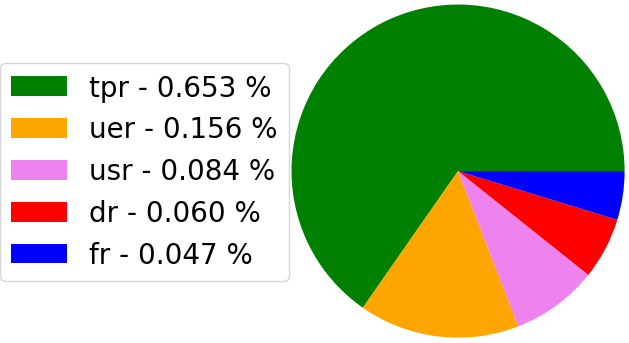} &
    \includegraphics[width=0.3333\textwidth]{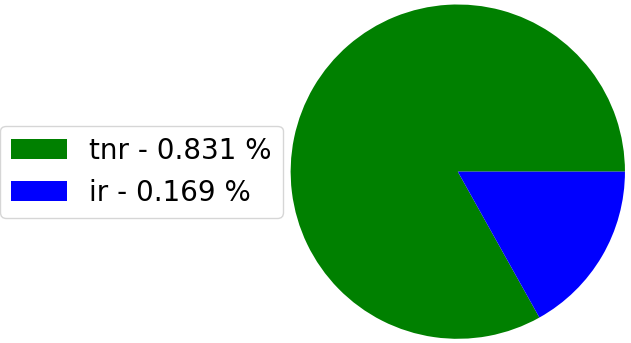} \\
    \raisebox{1\normalbaselineskip}[0pt][0pt]{\rotatebox{90}{COCO + neutral}} & 
    \includegraphics[width=0.3333\textwidth]{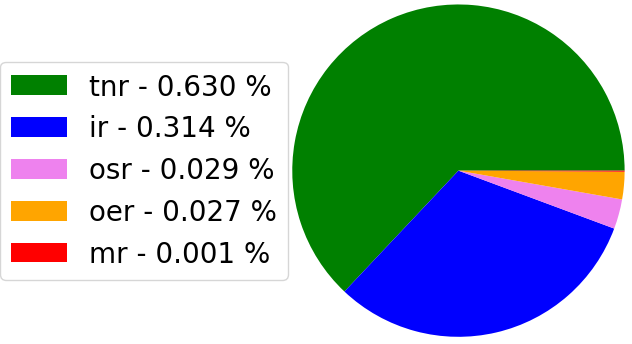} &
    \includegraphics[width=0.3333\textwidth]{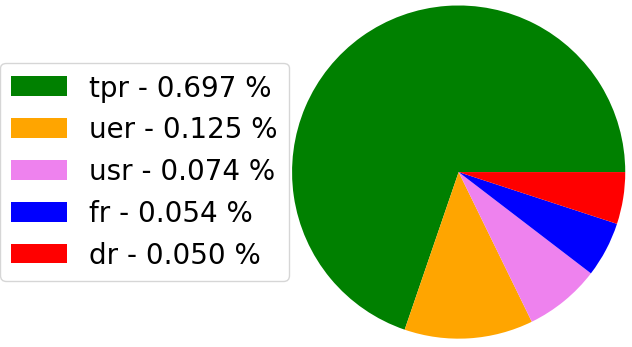} &
    \includegraphics[width=0.3333\textwidth]{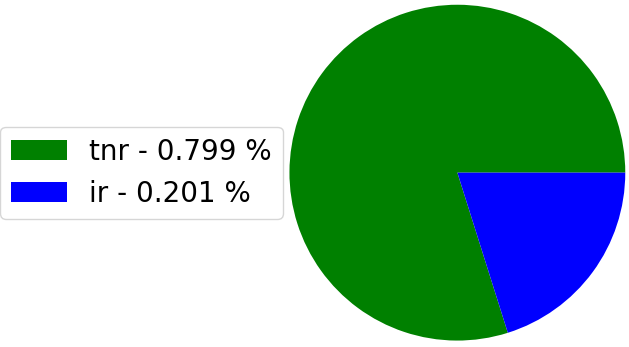} \\
    \raisebox{1.5\normalbaselineskip}[0pt][0pt]{\rotatebox{90}{Generated}} &
    \includegraphics[width=0.3333\textwidth]{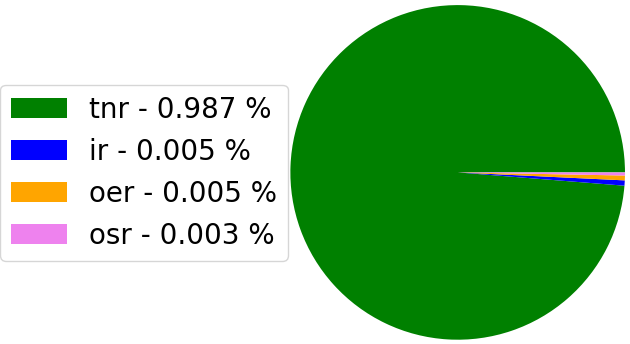} &
    \includegraphics[width=0.3333\textwidth]{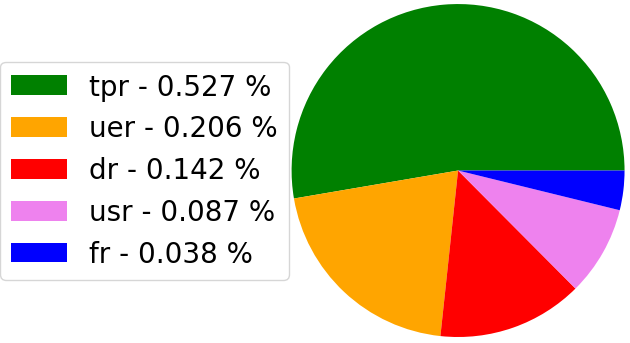} & 
    \includegraphics[width=0.3333\textwidth]{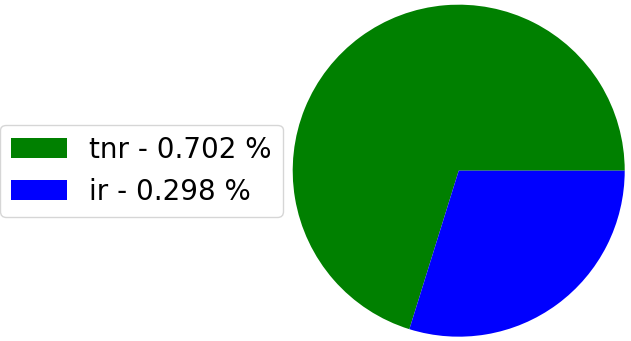} \\
    \raisebox{2\normalbaselineskip}[0pt][0pt]{\rotatebox{90}{Neutral}} & 
    \includegraphics[width=0.3333\textwidth]{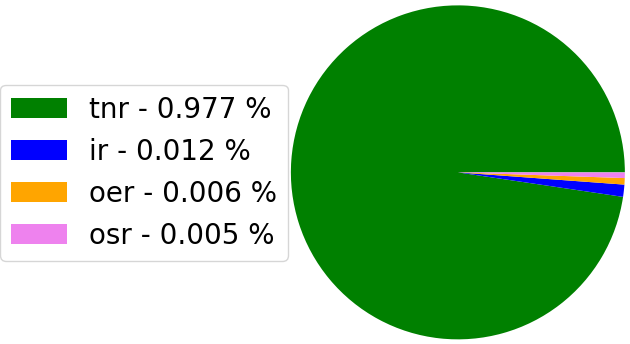} &
    \includegraphics[width=0.3333\textwidth]{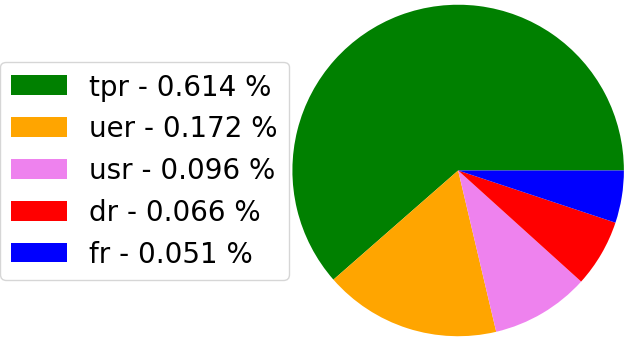} &
    \includegraphics[width=0.3333\textwidth]{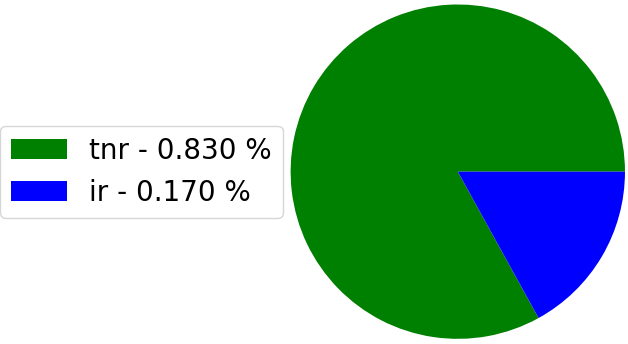}
\end{tabular}}
\label{fig:vanilla-generated-neutral}
\end{table*}

\subsection{Conclusion}

We deduce that the `\textit{vanilla}' approach is sensible to spurious detection. The `\textit{generated}' approach mitigates the problem, at the expense of detection performance.

\section{\uppercase{Handling neutral poses}}
\label{sec:neutral}

\subsection{Method proposed}
To get around this problem, we add `neutral' poses to the training set. We expect that this will improve the performance for two reasons. First, this action class frequently resets the system when the subject is not interacting with the robot, and thus avoid detecting many false-positives coming from aggregated noise. Indeed, when an action is detected, it clears the cost matrix, even though it does not allow simultaneous action (e.g. clapping while sitting). Second, this additional action class would help for the detection triggering issue. It forces the system to attribute negative score when the action is finished, because the user comes back to this initial pose, which corresponds or is contained in the `neutral' class. This implementation is referred as `\textit{neutral}'.

\subsection{Results}
The last line of the Figure~\ref{fig:vanilla-generated-neutral} presents the results obtained with the `\textit{neutral}' approach. This approach keeps the $tnr$ at a value as high as for the `\textit{vanilla}' approach tested on the 5 actions to detect, but also maintains a high value for the $tpr$ as well as the $tnr$ for neutral actions. 

While the `\textit{neutral}' approach seems to be the most trivial to handle `neutral' actions, it is the one that obtains the best results in our quantitative evaluation. It appears that the `neutral' actions are close enough between each others to be classified together and differentiated from the other actions. This approach allows performance equal to the `\textit{vanilla}' approach, while also handling real-case scenarios.

\subsection{Conclusion}

The three recognition methods presented in this article have been tested in our complete framework. While a rigorous quantitative assessment was not possible with this framework, our qualitative evaluation of the running system confirmed us that the `\textit{neutral}' approach is the best way to classify neutral poses.

%% file: sec_conclusion.tex
\section{\uppercase{conclusion}}
\label{sec:conclusion}

We have demonstrated an effective system for improving the human-Pepper interaction. Specifically, we have proposed a multi-step system composed of a body-pose estimator, a simple tracker and a gesture detection and recognition module on a super-charged Pepper robot. Our achievement was to run the whole pipeline with reasonable results in real-time despite the limited computational power available. Moreover, we noticed the drop of performance on real-case scenarios, and proposed a simple solution to overcome this problem. The resulting system is suitable for effective gesture recognition in real-world scenarios.

To improve each module independently and get even better performance, additional speed-up can be achieved by tweaking the Openpose architecture for mobile-friendly device (e.g. MobileNets \cite{mobilenets}) \cite{speed_accuracy_tradeoffs}, or by manually pruning filters \cite{pruning_2} to reduce the number of parameters and operations.

One recurring problem on gesture recognition is the lack of 3D skeletons, providing world-frame computation and invariance. A simple solution would be to actually estimate the 3D body-pose, by using depth sensors or networks able to infer the 3D location of the joint in the world frame \cite{densepose}.

%% file: main.bbl
\begin{thebibliography}{}

\bibitem[Andriluka et~al., 2014]{mpii}
Andriluka, M., Pishchulin, L., Gehler, P., and Schiele, B. (2014).
\newblock 2d human pose estimation: New benchmark and state of the art
  analysis.
\newblock In {\em IEEE Conference on Computer Vision and Pattern Recognition
  (CVPR)}.

\bibitem[Bentley, 1984]{kadane_algorithm}
Bentley, J. (1984).
\newblock Programming pearls: Algorithm design techniques.
\newblock {\em Commun. ACM}, 27(9):865--873.

\bibitem[Borghi et~al., 2016]{7899766}
Borghi, G., Vezzani, R., and Cucchiara, R. (2016).
\newblock Fast gesture recognition with multiple stream discrete hmms on 3d
  skeletons.
\newblock In {\em 23rd International Conference on Pattern Recognition (ICPR)}.

\bibitem[Cao et~al., 2016]{openpose}
Cao, Z., Simon, T., Wei, S., and Sheikh, Y. (2016).
\newblock Realtime multi-person 2d pose estimation using part affinity fields.
\newblock arXiv 1611.08050.

\bibitem[Du et~al., 2015]{du_hierarchical}
Du, Y., Wang, W., and Wang, L. (2015).
\newblock Hierarchical recurrent neural network for skeleton based action
  recognition.
\newblock In {\em 2015 {IEEE} {Conference} on {Computer} {Vision} and {Pattern}
  {Recognition} ({CVPR})}, pages 1110--1118.

\bibitem[Ellis et~al., 2013]{Ellis:2013:ETA:2440638.2440682}
Ellis, C., Masood, S.~Z., Tappen, M.~F., Laviola, Jr., J.~J., and Sukthankar,
  R. (2013).
\newblock Exploring the trade-off between accuracy and observational latency in
  action recognition.
\newblock {\em Int. J. Comput. Vision}, 101(3):420--436.

\bibitem[Evangelidis et~al., 2014]{Evangelidis-ECCVW-2014}
Evangelidis, G., Singh, G., and Horaud, R. (2014).
\newblock Continuous gesture recognition from articulated poses.
\newblock In {\em ECCV Workshops}.

\bibitem[Fan et~al., 2012]{6376770}
Fan, Z., Li, G., Haixian, L., Shu, G., and Jinkui, L. (2012).
\newblock Star skeleton for human behavior recognition.
\newblock In {\em 2012 International Conference on Audio, Language and Image
  Processing}, pages 1046--1050.

\bibitem[G{\"{u}}ler et~al., 2018]{densepose}
G{\"{u}}ler, R.~A., Neverova, N., and Kokkinos, I. (2018).
\newblock Densepose: Dense human pose estimation in the wild.
\newblock arXiv 1802.00434.

\bibitem[Han et~al., 2016]{gesture_review_1}
Han, F., Reily, B., Hoff, W., and Zhang, H. (2016).
\newblock Space-time representation of people based on 3d skeletal data: {A}
  review.
\newblock arXiv 1601.01006.

\bibitem[He et~al., 2017]{mask_rcnn}
He, K., Gkioxari, G., Doll{\'{a}}r, P., and Girshick, R.~B. (2017).
\newblock Mask {R-CNN}.
\newblock arXiv 1703.06870.

\bibitem[Howard et~al., 2017]{mobilenets}
Howard, A.~G., Zhu, M., Chen, B., Kalenichenko, D., Wang, W., Weyand, T.,
  Andreetto, M., and Adam, H. (2017).
\newblock Mobilenets: Efficient convolutional neural networks for mobile vision
  applications.
\newblock arXiv 1704.04861.

\bibitem[Huang et~al., 2016]{speed_accuracy_tradeoffs}
Huang, J., Rathod, V., Sun, C., Zhu, M., Korattikara, A., Fathi, A., Fischer,
  I., Wojna, Z., Song, Y., Guadarrama, S., and Murphy, K. (2016).
\newblock Speed/accuracy trade-offs for modern convolutional object detectors.
\newblock arXiv 1611.10012.

\bibitem[Huang et~al., 2018]{pruning_2}
Huang, Q., Zhou, S.~K., You, S., and Neumann, U. (2018).
\newblock Learning to prune filters in convolutional neural networks.
\newblock arXiv 1801.07365.

\bibitem[Insafutdinov et~al., 2016a]{arttrack}
Insafutdinov, E., Andriluka, M., Pishchulin, L., Tang, S., Levinkov, E.,
  Andres, B., and Schiele, B. (2016a).
\newblock Articulated multi-person tracking in the wild.
\newblock arXiv 1612.01465.

\bibitem[Insafutdinov et~al., 2016b]{deepercut}
Insafutdinov, E., Pishchulin, L., Andres, B., Andriluka, M., and Schiele, B.
  (2016b).
\newblock Deepercut: {A} deeper, stronger, and faster multi-person pose
  estimation model.
\newblock arXiv 1605.03170.

\bibitem[Jin and Choi, 2013]{orientation_feature_1}
Jin, S.-Y. and Choi, H.-J. (2013).
\newblock Essential body-joint and atomic action detection for human activity
  recognition using longest common subsequence algorithm.
\newblock In {\em Computer Vision - ACCV 2012 Workshops}, pages 148--159.

\bibitem[Li et~al., 2016]{li2016online}
Li, Y., Lan, C., Xing, J., Zeng, W., Yuan, C., and Liu, J. (2016).
\newblock Online human action detection using joint classification-regression
  recurrent neural networks.
\newblock {\em European Conference on Computer Vision}.

\bibitem[Lin et~al., 2014]{coco_dataset}
Lin, T., Maire, M., Belongie, S.~J., Bourdev, L.~D., Girshick, R.~B., Hays, J.,
  Perona, P., Ramanan, D., Doll{\'{a}}r, P., and Zitnick, C.~L. (2014).
\newblock Microsoft {COCO:} common objects in context.
\newblock arXiv 1405.0312.

\bibitem[Liu et~al., 2017a]{7952447}
Liu, C., Li, Y., Hu, Y., and Liu, J. (2017a).
\newblock Online action detection and forecast via multitask deep recurrent
  neural networks.
\newblock In {\em 2017 IEEE International Conference on Acoustics, Speech and
  Signal Processing (ICASSP)}, pages 1702--1706.

\bibitem[Liu et~al., 2017b]{attention_lstm}
Liu, J., Wang, G., Duan, L., Hu, P., and Kot, A.~C. (2017b).
\newblock Skeleton based human action recognition with global context-aware
  attention {LSTM} networks.
\newblock arXiv 1707.05740.

\bibitem[Luo et~al., 2013]{displacement_feature}
Luo, J., Wang, W., and Qi, H. (2013).
\newblock Group sparsity and geometry constrained dictionary learning for
  action recognition from depth maps.
\newblock In {\em 2013 IEEE International Conference on Computer Vision}.

\bibitem[Meshry et~al., 2016]{DBLP:journals/corr/MeshryHT15}
Meshry, M., Hussein, M.~E., and Torki, M. (2016).
\newblock Linear-time online action detection from 3d skeletal data using bags
  of gesturelets.
\newblock {\em 2016 IEEE Winter Conference on Applications of Computer Vision
  (WACV)}.

\bibitem[Newell et~al., 2016]{stacked_hourglass}
Newell, A., Yang, K., and Deng, J. (2016).
\newblock Stacked hourglass networks for human pose estimation.
\newblock arXiv 1603.06937.

\bibitem[Nowozin and Shotton, 2012]{shotton}
Nowozin, S. and Shotton, J. (2012).
\newblock Action points: A representation for low-latency online human action
  recognition.
\newblock Technical report.

\bibitem[Papandreou et~al., 2017]{Papandreou}
Papandreou, G., Zhu, T., Kanazawa, N., Toshev, A., Tompson, J., Bregler, C.,
  and Murphy, K.~P. (2017).
\newblock Towards accurate multi-person pose estimation in the wild.
\newblock arXiv 1701.01779.

\bibitem[Pham et~al., 2018]{PHAM2018}
Pham, H.-H., Khoudour, L., Crouzil, A., Zegers, P., and Velastin, S.~A. (2018).
\newblock Exploiting deep residual networks for human action recognition from
  skeletal data.
\newblock {\em Computer Vision and Image Understanding}.

\bibitem[Piyathilaka and Kodagoda, 2013]{6566433}
Piyathilaka, L. and Kodagoda, S. (2013).
\newblock Gaussian mixture based hmm for human daily activity recognition using
  3d skeleton features.
\newblock In {\em 2013 IEEE 8th Conference on Industrial Electronics and
  Applications (ICIEA)}.

\bibitem[Presti and Cascia, 2016]{gesture_review_2}
Presti, L.~L. and Cascia, M.~L. (2016).
\newblock 3d skeleton-based human action classification: A survey.
\newblock {\em Pattern Recognition}, 53:130 -- 147.

\bibitem[Russakovsky et~al., 2014]{imagenet}
Russakovsky, O., Deng, J., Su, H., Krause, J., Satheesh, S., Ma, S., Huang, Z.,
  Karpathy, A., Khosla, A., Bernstein, M.~S., Berg, A.~C., and Li, F. (2014).
\newblock Imagenet large scale visual recognition challenge.
\newblock arXiv 1409.0575.

\bibitem[Shahroudy et~al., 2016]{DBLP:journals/corr/ShahroudyLNW16}
Shahroudy, A., Liu, J., Ng, T., and Wang, G. (2016).
\newblock {NTU} {RGB+D:} {A} large scale dataset for 3d human activity
  analysis.
\newblock arXiv 1604.02808.

\bibitem[Slama et~al., 2015]{grassmann_manifold}
Slama, R., Wannous, H., Daoudi, M., and Srivastava, A. (2015).
\newblock Accurate 3d action recognition using learning on the grassmann
  manifold.

\bibitem[Tompson et~al., 2014]{confidence_heatmap}
Tompson, J., Jain, A., LeCun, Y., and Bregler, C. (2014).
\newblock Joint training of a convolutional network and a graphical model for
  human pose estimation.
\newblock arXiv 1406.2984.

\bibitem[Toshev and Szegedy, 2013]{deeppose}
Toshev, A. and Szegedy, C. (2013).
\newblock Deeppose: Human pose estimation via deep neural networks.
\newblock arXiv 1312.4659.

\bibitem[Vemulapalli et~al., 2014]{lie_manifold}
Vemulapalli, R., Arrate, F., and Chellappa, R. (2014).
\newblock Human action recognition by representing 3d skeletons as points in a
  lie group.
\newblock In {\em Proceedings of the 2014 IEEE Conference on Computer Vision
  and Pattern Recognition}, CVPR '14, pages 588--595.

\bibitem[Ward et~al., 2011]{metrics_activity_reco}
Ward, J.~A., Lukowicz, P., and Gellersen, H.~W. (2011).
\newblock Performance metrics for activity recognition.
\newblock {\em ACM Trans. Intell. Syst. Technol.}, 2(1):6:1--6:23.

\bibitem[Wei et~al., 2016]{cpm}
Wei, S., Ramakrishna, V., Kanade, T., and Sheikh, Y. (2016).
\newblock Convolutional pose machines.
\newblock arXiv 1602.00134.

\bibitem[Xia et~al., 2012]{orientation_feature_3}
Xia, L., Chen, C.~C., and Aggarwal, J.~K. (2012).
\newblock View invariant human action recognition using histograms of 3d
  joints.
\newblock In {\em 2012 IEEE Computer Society Conference on Computer Vision and
  Pattern Recognition Workshops}, pages 20--27.

\bibitem[Yan et~al., 2018]{graph_skeleton}
Yan, S., Xiong, Y., and Lin, D. (2018).
\newblock Spatial temporal graph convolutional networks for skeleton-based
  action recognition.
\newblock arXiv 1801.07455.

\bibitem[Yin, 2014]{Yin2014RealtimeCG}
Yin, Y. (2014).
\newblock Real-time continuous gesture recognition for natural multimodal
  interaction.

\bibitem[Zanfir et~al., 2013]{moving_pose}
Zanfir, M., Leordeanu, M., and Sminchisescu, C. (2013).
\newblock The moving pose: An efficient 3d kinematics descriptor for
  low-latency action recognition and detection.
\newblock In {\em 2013 IEEE International Conference on Computer Vision}, pages
  2752--2759.

\end{thebibliography}
